\documentclass[copyright,creativecommons]{eptcs}

\usepackage{subcaption}
\usepackage{graphicx}
\usepackage{pgffor}

\usepackage{nameref}
\makeatletter
\newcommand*{\currentname}{\@currentlabelname}
\makeatother

\usepackage[disable]{todonotes}
\hypersetup{hidelinks}
\usepackage{iftex}
\ifpdf
  \usepackage{underscore}         
  \usepackage[T1]{fontenc}        
\else
  \usepackage{breakurl}           
\fi

\usepackage{acronym}
\newacro{AVV}{Armoured Vanguard Vehicle}
\newacro{UAV}{Unmanned Aerial Vehicle}
\newacro{UGV}{Unmanned Ground Vehicle}
\newacro{HTN}{Hierarchical Task Network}
\newacro{HMI}{Human-Machine Interface}

\usepackage{forest}
\usepackage{tikz}
\usetikzlibrary{shapes}
\tikzstyle{disp}   = [draw, align=center]
\tikzstyle{action} = [disp, rectangle, rounded corners]
\tikzstyle{method} = [disp, signal, signal to=west and east]
\tikzstyle{task}   = [disp, ellipse, inner sep=.05cm]

\newenvironment{chronicle}{
    \begin{center}
    \begin{tabular}[htbp]{l@{}l}
}{
    \end{tabular}
    \end{center}
}
\newcommand{\cname}[1]{\\\multicolumn{2}{@{}l}{\textsf{#1}}}
\newcommand{\ckey}[1]{\\\texttt{#1}:&}
\newcommand{\cnl}{\\&}

\usepackage[inline]{enumitem}
\newlist{plan}{itemize}{1}
\setlist[plan]{label=, itemsep=0ex}
\newcommand\paction[0]{\item}

\newlist{ilist}{enumerate*}{1}
\setlist[ilist]{label=(\roman*)}

\def\eg{\textit{e.g.}~}
\def\ie{\textit{i.e.}~}

\def\params{\tuple{x_1, \dots, x_n}}

\newcommand{\set}[1]{\left\{\:#1\:\right\}}
\newcommand{\tuple}[1]{\left(\:#1\:\right)}

\def\titlerunning{Multi-Robot Task Planning to Secure Human Group Progress}
\def\authorrunning{R. Godet, C. Lesire, A. Bit-Monnot}

\title{\titlerunning}

\author{Roland Godet\textsuperscript{\rm 1,2}
\email{roland.godet@laas.fr}
\and Charles Lesire\textsuperscript{\rm 1}
\email{charles.lesire@onera.fr}
\and Arthur Bit-Monnot\textsuperscript{\rm 2}
\email{arthur.bit-monnot@laas.fr}
\institute{\textsuperscript{\rm 1} ONERA/DTIS, University of Toulouse, France}
\institute{\textsuperscript{\rm 2} LAAS-CNRS, University of Toulouse, INSA, Toulouse, France}
}

\begin{document}
\maketitle


\begin{abstract}
    Recent years have seen an increasing number of deployment of fleets of autonomous vehicles.
    As the problem scales up, in terms of autonomous vehicles number and complexity of their objectives, there is a growing need for decision-support tooling to help the operators in controlling the fleet.

    In this paper, we present an automated planning system developed to assist the operators in the CoHoMa II challenge, where a fleet of robots, remotely controlled by a handful of operators, must explore and progress through a potential hostile area.
    In this context, we use planning to provide the operators with suggestions about the actions to consider and their allocation to the robots.

    This paper especially focus on the modelling of the problem as a hierarchical planning problem for which we use a state-of-the-art automated solver.
\end{abstract}


\section{Introduction}

The "Battle-Lab Terre", a part of the French Army studying innovation, organized in 2022 the second version of the CoHoMa challenge~\cite{noauthor:challenge:2022} in order to study the collaboration between human operators and autonomous multi-robot systems.

\begin{figure}[htbp]
    \centering
    \begin{subfigure}{0.37\linewidth}
        \centering
        \includegraphics[width=\linewidth]{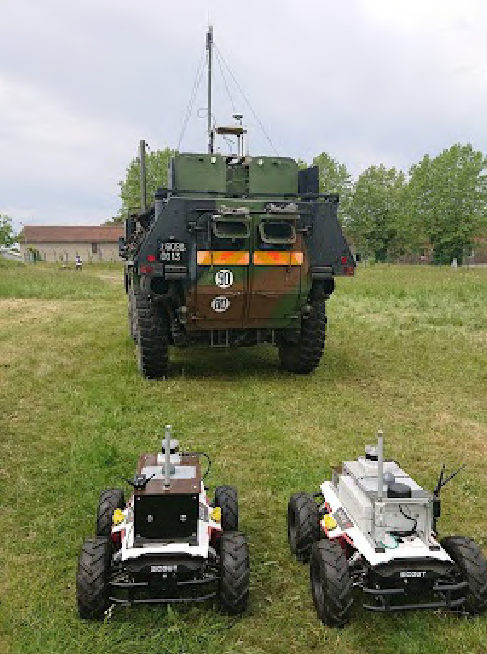}
        \caption{The \ac{AVV} and two \acp{UGV}}
        \label{fig:avv-ugv}
    \end{subfigure}
    \hfill
    \begin{subfigure}{0.6\linewidth}
        \centering
        \includegraphics[width=\linewidth]{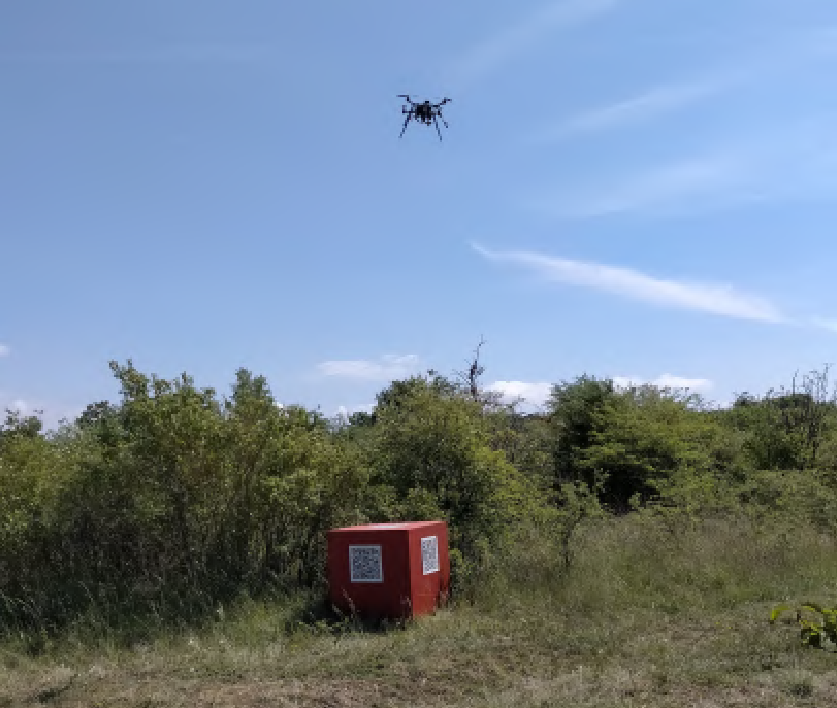}
        \caption{A \ac{UAV} detecting a trap}
        \label{fig:red-cube}
    \end{subfigure}
    \caption{Illustrations of the CoHoMa challenge}
\end{figure}

The task was to navigate through a dangerous terrain in an \ac{AVV} (\autoref{fig:avv-ugv}).
The land included 1m-wide red cube (\autoref{fig:red-cube}) representing a trap said to be explosive and capable of damaging the \ac{AVV}.
Therefore, the human operators on board had to ensure that the \ac{AVV}'s environment was safe before moving it.
To do this, they had to use various \acp{UAV} and \acp{UGV}, to perform reconnaissance missions, seek out traps, and avoid or disable them.
A general system architecture of these vehicles has been studied in~\cite{flecher:imugs:2022}.

When the number of unmanned vehicles is too important for the number of human operators (6 \acp{UGV} and 3 \acp{UAV} for 4 human operators in our case), a decision-making aid is welcomed.
This aid must decide which actions are to be performed, when, and by which vehicle.
This problem of multi-robot task allocation is highly studied~\cite{gini:multi-robot:2017}, especially when there are communications issues~\cite{bechon:planification:2016} which will be ignored in this study.

The model proposed in this paper is rooted in the CoHoMa challenge. At a high level it abstracts of emergency and rescue missions~\cite{delmerico:current:2019} such as floods controlling~\cite{tang:temporal:2013} or subterranean rescue~\cite{orekhov:darpa:2022}, using mixed-initiative planning with automated vehicles~\cite{bevacqua:mixed-initiative:2015}.

The mission is for a group of humans to go through a hazardous zone with securable obstacles that they must avoid.
Because the obstacles are unknown at the beginning of the mission, the operators have at their disposal \acp{UAV} and \acp{UGV} to explore the area, detect obstacles and secure them.
The fleet of robots is typically heterogeneous: they have different capacities, in order to be complementary and be able to secure the human movements.
The obstacles will be discovered as the progression goes on, so there will be replanning steps for each event.

\begin{figure}[htbp]
    \centering
    \includegraphics[width=0.6\linewidth]{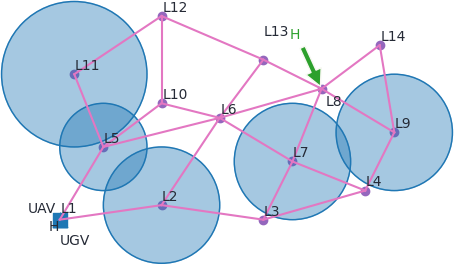}
    \caption[]{
        \centering
        Example of a navigation graph where (i) the vehicles (\ac{UAV}, \ac{UGV}, H for Humans) are in L1 (ii) the objective is for H to go to L8 (iii) there are undetected obstacles in L2, L5, L6, L9 and L11
    }
    \label{fig:mission-0}
\end{figure}

To simplify the interactions with the robots, their locations are discretized.
Indeed, the operator does not need to have a precise representation of the robot's location for the planning process, the points of interest are sufficient.
Therefore, a navigation graph as shown in the \autoref{fig:mission-0} is used.
This graph regroups the location of the vehicles, the location of the obstacles, and the objectives of the mission.
Moreover, the edges of the graph are configured to forbid the access to some vehicles, \eg a \ac{UAV} can cross a cliff where the other vehicles cannot.
This way, a unique graph can be used to store all the possible displacements.

\begin{figure}[htbp]
    \centering
    \begin{subfigure}{0.59\linewidth}
        \centering
        \includegraphics[width=\linewidth]{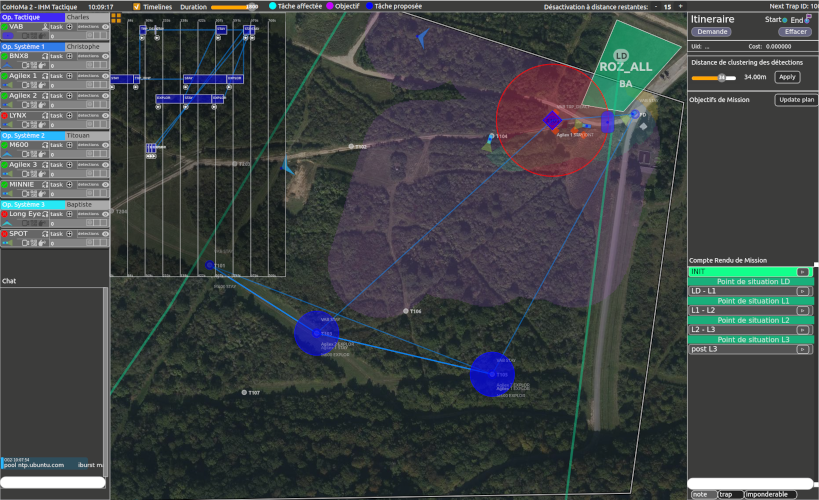}
        \caption{Full view of the \ac{HMI}}
        \label{fig:hmi-full}
    \end{subfigure}
    \hfill
    \begin{subfigure}{0.4\linewidth}
        \centering
        \includegraphics[width=\linewidth]{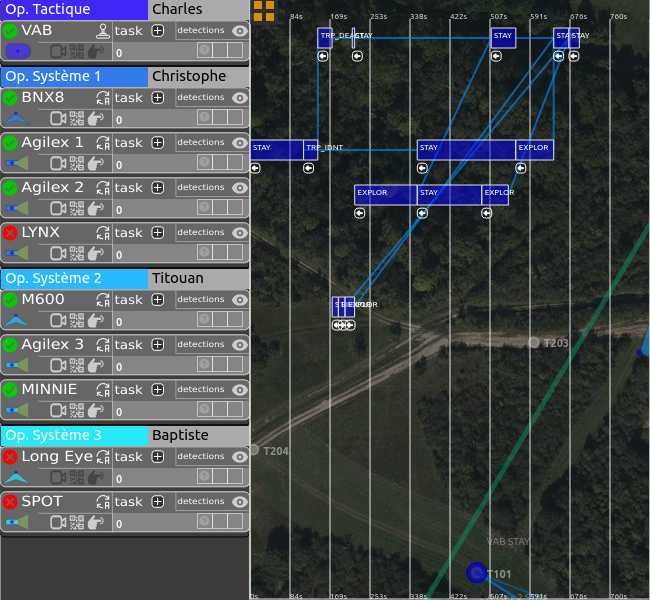}
        \caption{Zoom on the plan visual: a line is the timeline of a robot}
        \label{fig:hmi-zoom}
    \end{subfigure}
    \caption{\ac{HMI} used by the operator to interact with the robot fleet}
\end{figure}

The \autoref{fig:hmi-full} shows the \ac{HMI} used by a human operator to visualize the environment, the real location of the robots and the detected obstacles, \ie the navigation graph with more details, on a satellite view of the terrain.
The operator can interact with the map to specify events, \eg an obstacle detection, and to change the mission's objectives.
When the mission details have been updated, the operator can request a plan to achieve those objectives on the right side of the \ac{HMI}.
This plan is not sent to the robots directly.
First, it is shown in the \ac{HMI} (see \autoref{fig:hmi-zoom}) on the left side for potential modification, \eg allocate an action to another robot, and for approbation.
Thus, the plan needs to be as simple as possible in order to be easily understandable by the operator.
Once the plan is validated, it is sent to each robot which are able to accomplish it.
For example, considering the calculated plan
\begin{plan}
    \paction Move \ac{UAV} from $L_1$ to $L_5$
    \paction Move \ac{UAV} from $L_5$ to $L_{10}$
    \paction Move \ac{UAV} from $L_{10}$ to $L_{12}$
\end{plan}
\noindent
The operator does not need to know which path the vehicle will take since it is autonomous, so the tasks can be regrouped into a single task 'Move \ac{UAV} from $L_1$ to $L_{12}$'.
Eventually, the operator already knows where the UAV is at the beginning (\ie in location $L_1$), so the action can be transformed, and the plan can be simplified as 'Move \ac{UAV} to $L_{12}$'.
After validation, the task is sent to the concerned robot \ac{UAV}, which knows how to go to the location $L_{12}$.

Although the robots are capable of detecting obstacles, they do not modify the mission on their own because their detection cannot be perfectly accurate; they add several false obstacles next to the real one.
To compensate for this, the detected obstacles are displayed in the \ac{HMI} by grouping nearby obstacles together, with a customizable threshold, and operator approval is required to add the obstacle to the mission problem.
This approach is generalized to all possible events.
In this way, no uncertainty is taken into account in the planning process; it is taken upstream by the operator who has validated the event.
Finally, as two events can occur at the same time for two different robots, replanning is not triggered automatically after each event but only when the operator requests it.

This paper will begin by presenting the necessary background for chronicle modelling.
Next, a model that is as simple as possible for a non-expert user, called the \textit{natural} model in the following, will be proposed to show the limitations of simple models.
Finally, some optimizations of this first model will be introduced, and the time needed for the planner to find a solution will be compared.


\section{Background}

To model the planning problem, we wish to exploit the hierarchical nature of the task where some high-level tasks to accomplished are specified by the operator that must then be refined into sets of primitives actions executable by the autonomous vehicles.

An \ac{HTN}\cite{bercher:survey:2019} can represent this kind of decomposition and is easily defined with the HDDL language~\cite{holler:hddl:2020}.
This language however lacks the ability to express temporal properties of the problem such as the duration of action or deadlines.
Instead, we rely on the formalism of chronicles~\cite{ghallab:automated:2004} that support the specification of rich temporal planning problem. In particular, we exploit their extension for hierarchical task networks can represent combined temporal and hierarchical problems~\cite{godet:chronicles:2022}. However, it does not have an input language that can represent both.

A \textbf{type} is a set of values that can be either domain constants (\eg the type $Vehicle=\set{V_1, V_2}$ defines two vehicles objects $V_1$, $V_2$) or numeric values (\eg \textbf{timepoints} are regularly spaced numerical values describing absolute times when events occur).
The types can present a hierarchy, \eg the type $Robot$ is a subtype of $Vehicle$ meaning that a $Robot$ is a $Vehicle$, but the reverse is not necessarily true.
When there is a type hierarchy, an abstract root type named $Object$ is defined in order to have a decomposition tree.

A \textbf{state variable} describes the evolution of an environment characteristic over time.
Generally, it is parametrized by one or multiple variables.
Its value will depend on the value of the variables, \eg $loc(v)$ denotes the evolution of the location of the vehicle $v$, its value will be $loc(V_1)$ or $loc(V_2)$ depending on the value taken by $v$ of type $Vehicle$.

A \textbf{task} is a high-level operation to accomplish over time.
Generally, it is parametrized by one or multiple variables.
It is of the form $[s,e]task\params$ where $s$ and $e$ are timepoints denoting the start and end instants when the task occurs, $task\params$ is the task with each $x_i$ a variable.
For instance, $[2,4]Move(V_1, L_2)$ denotes the operation of moving the vehicle $V_1$ to the location $L_2$ during the temporal interval $[2,4]$.
The set of available tasks of the planning problems is $\mathcal{T}$.

A \textbf{chronicle} defines the requirements of a process in the planning problem.
A chronicle is a tuple $\mathcal{C}=\tuple{V,T,X,C,E,S}$ where:
\begin{itemize}
    \item $V$ is the set of \textit{variables} of the chronicle. This set is split into a set of temporal variables $V_T$ whose domains are timepoints and a set of non-temporal variables $V_O$.

    \item $T\in \mathcal{T}$ is the parametrized \textit{task} achieved by the chronicle. The start and the end instants of the task correspond to the start and the end instants when the chronicle is active, it is its \textit{active temporal interval}.

    \item $X$ is a set of \textit{constraints} over the variables of $V$. The chronicle cannot be \textit{active} (defined bellow) if at leat one constraint is not respected over its active temporal interval.

    \item $C$ is a set of \textit{conditions} with each condition of the form $[s,e]var\params=v$ where $\tuple{s,e}\in V_T^2$ such that the temporal interval $[s,e]$ is contained in the active temporal interval of the chronicle, $var\params$ is a parametrized state variable with each $x_i\in V_O$, and $v\in V_O$. A condition is verified if the state variable $var\params$ has the value $v$ over the temporal interval $[s,e]$. The chronicle cannot be active if at least one condition is not verified.

    \item $E$ is a set of \textit{effects} with each effect of the form $[s,e]var\params\gets v$ where $\tuple{s,e}\in V_T^2$ such that the temporal interval $[s,e]$ is contained in the active temporal interval of the chronicle, $var\params$ is a parametrized state variable with each $x_i\in V_O$, and $v\in V_O$. An effect states that the state variable $var\params$ takes the value $v$ at time $e$. The temporal interval $]s,e[$ is the moment when the state variable is transitioning from its previous value to its new value. During this transition, the value of the state variable is undetermined.

    \item $S$ is a set of \textit{subtasks} where each subtask is a task in $\mathcal{T}$ that must be achieved by another chronicle.
\end{itemize}

\noindent
A chronicle can be \textbf{active} or not, defining whether the chronicle is present in the final solution.
If the chronicle is not active, then the planner must find another chronicle achieving the same task to replace it.

We make the distinction between three types of chronicles: the \textbf{action chronicle} which has effects but no subtasks (\ie $S=\emptyset$), the \textbf{method chronicle} which has subtasks but no effects (\ie $E=\emptyset$), and the \textbf{initial chronicle} encoding the initial state as effect and the objectives of the problem as conditions and subtasks, it is the only one which does not have a task $T$ to achieve (\ie $T=\emptyset$).

As an alternative to specifying chronicles manually, the AIPlan4EU project\footnote{\url{https://www.aiplan4eu-project.eu/}} offers a Python API\footnote{\url{https://github.com/aiplan4eu/unified-planning}} for modelling different kinds of planning problems, notably temporal and hierarchical ones. The corresponding problems map almost immediately to the chronicles defined above.
The python API for constructing planning problems is especially useful in our case where the new problems are defined online, as the situation evolves during the mission.


\section{Initial Model}

According to the mission specification, the humans need to be able to move while the autonomous vehicles need to move, explore to detect obstacles and secure them.
This way, a list of high-level tasks appears:
\begin{itemize}
    \item $[s,e]goto(v, l)$ : The vehicle $v$ (humans, \ac{UAV} or \ac{UGV}) goes to the location $l$
    \item $[s,e]explore(r, f, t)$: The robot $r$ (\ac{UAV} or \ac{UGV}) explores the path from the location $f$ to $t$
    \item $[s,e]secure(r, o)$: The robot $r$ secures the obstacle $o$
\end{itemize}

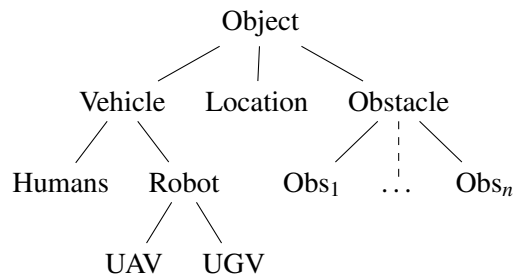
\begin{figure}[htbp]
    \centering
    \begin{forest}
 [Object
   [Vehicle
     [Humans]
     [Robot
       [UAV]
       [UGV]
     ]
   ]
   [Location]
   [Obstacle
     [Obs$_1$]
     [\dots, edge=dashed]
     [Obs$_n$]
   ]
 ]
\end{forest}
    \caption{Type hierarchy}
    \label{fig:type-hierarchy}
\end{figure}

\noindent
From this list, one can easily extract the type hierarchy shown in the \autoref{fig:type-hierarchy}.
The \textit{Obstacle} allows handling different types of obstacles for the $secure$ task, \eg in a fire rescue mission we could image to use different types of extinguishers (water, CO$_2$ or powder), each one for a different type of obstacle.

\subsection{Goto task}

A vehicle needs to be able to go from a location to another.
However, a human, a \ac{UAV} and a \ac{UGV} does not move the same way.
A human will \textit{walk} while a \ac{UAV} will \textit{fly} and a \ac{UGV} will \textit{roll} on land. Therefore, we obtain the three following action chronicles:

\begin{minipage}{0.5\textwidth}
    \begin{chronicle}
        \cname{$[s,e]walk(h, f, t)$}
        \ckey{variables} Humans $h$
        \cnl Locations $f$ (from) and $t$ (to)
        \ckey{task} $[s,e]walk(h, f, t)$
        \ckey{constraints} $f \ne t$
        \cnl $e-s=dur(h,f,t)$
        \ckey{conditions} $[s,s]loc(h)=f$
        \cnl $[s,e]path(f, t)=\top$
        \cnl $[s,e]explored\ air(f, t)=\top$
        \cnl $[s,e]explored\ ground(f, t)=\top$
        \cnl $[s,e]obstacle(f, t)=\bot$
        \ckey{effects} $[s,e]loc(h)\gets t$
    \end{chronicle}
\end{minipage}
\begin{minipage}{0.45\textwidth}
    The humans $h$ can move to the location $t$ only if
    \begin{ilist}
        \item the humans are in the location $f$ at the beginning of the chronicle
        \item there is an edge from $f$ to $t$ in the navigation graph
        \item the path has been explored by a \ac{UAV} and a \ac{UGV}
        \item there is no obstacle affecting the path.
    \end{ilist}

    At the end of the chronicle, the humans $h$ will be at the location $t$.

    The state variable $dur(v, f, t)$ represents the duration taken by the vehicle $v$ to go from the location $f$ to the location $t$.
    It depends on the distance between $f$ and $t$, and on the speed of the vehicle $v$.
\end{minipage}

\begin{minipage}{0.5\textwidth}
    \begin{chronicle}
        \cname{$[s,e]fly(a, f, t)$}
        \ckey{variables} \ac{UAV} $a$
        \cnl Locations $f$ (from) and $t$ (to)
        \ckey{task} $[s,e]fly(a, f, t)$
        \ckey{constraints} $f \ne t$
        \cnl $e-s=dur(a,f,t)$
        \ckey{conditions} $[s,s]loc(a)=f$
        \cnl $[s,e]path(f, t)=\top$
        \ckey{effects} $[s,e]loc(a)\gets t$
    \end{chronicle}
\end{minipage}
\begin{minipage}{0.45\textwidth}
    The \ac{UAV} $a$ can move to the location $t$ only if
    \begin{ilist}
        \item the \ac{UAV} is in the location $f$ at the beginning of the chronicle
        \item there is an edge from $f$ to $t$ in the navigation graph
    \end{ilist}

    At the end of the chronicle, the \ac{UAV} $a$ will be at the location $t$.

    As for the \textit{walk} chronicle, the duration is specified with the constraint $e-s=dur(a,f,t)$.
\end{minipage}

\noindent \\
The chronicle $[s,e]roll(g, f, t)$ is similar to the chronicle $[s,e]fly(a, f, t)$ by replacing the \ac{UAV} $a$ by the \ac{UGV} $g$.
However, the distinction is made because in a more detailed model it could be more conditions and effects making a difference between the air and ground movements.

\begin{figure}[htbp]
    \centering
    \begin{forest}
  [{goto(v, t)}, task
      [{Noop(v, t)}, method]
      [{UAV(v, f, i, t)}, method
          [{fly(v, f, i)}, action]
          [{goto(v, t)}, task]
      ]
      [{UGV(v, f, i, t)}, method
          [{roll(v, f, i)}, action]
          [{goto(v, t)}, task]
      ]
      [{Humans(v, f, i, t)}, method
          [{walk(v, f, i)}, action]
          [{goto(v, t)}, task]
      ]
  ]
\end{forest}
    \caption{Natural decomposition of the \textsf{goto} task where (i) rectangles are action chronicles (ii) diamonds are method chronicles and (iii) ovals are tasks to achieve}
    \label{fig:goto-naive}
\end{figure}
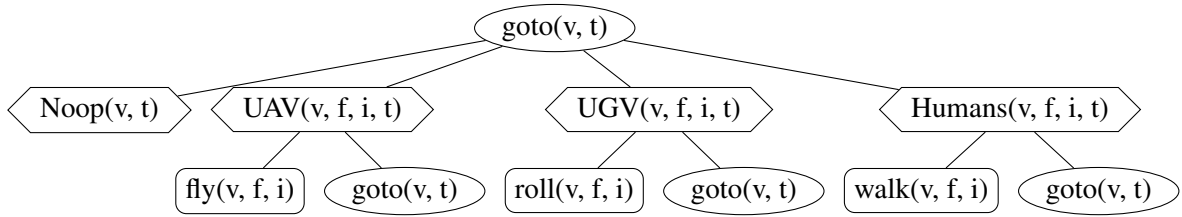

\begin{figure}[htbp]
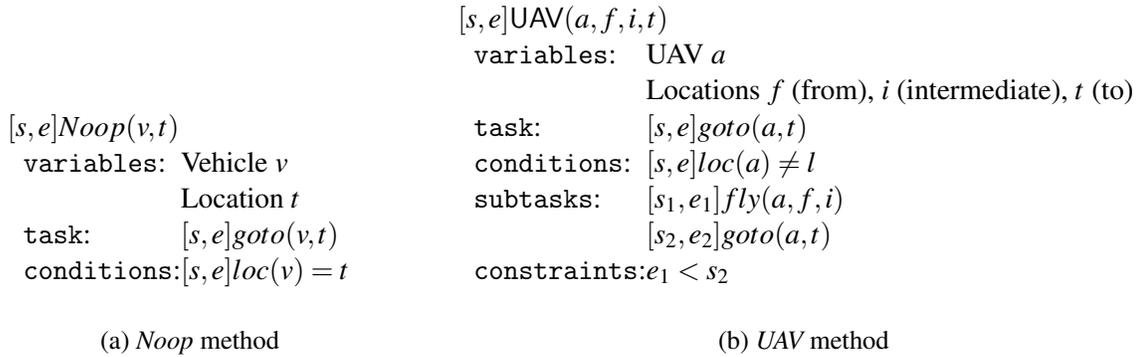

    \centering
    \begin{subfigure}{0.4\linewidth}
        \begin{chronicle}
            \cname{$[s,e]Noop(v, t)$}
            \ckey{variables} Vehicle $v$
            \cnl Location $t$
            \ckey{task} $[s,e]goto(v, t)$
            \ckey{conditions} $[s,e]loc(v)=t$
        \end{chronicle}
        \caption{\textit{Noop} method}
        \label{fig:goto-methods-noop}
    \end{subfigure}
    \hfill
    \begin{subfigure}{0.58\linewidth}
        \begin{chronicle}
            \cname{$[s,e]\ac{UAV}(a, f, i, t)$}
            \ckey{variables} \ac{UAV} $a$
            \cnl Locations $f$ (from), $i$ (intermediate), $t$ (to)
            \ckey{task} $[s,e]goto(a, t)$
            \ckey{conditions} $[s,e]loc(a)\ne l$
            \ckey{subtasks} $[s_1, e_1]fly(a, f, i)$
            \cnl $[s_2, e_2]goto(a, t)$
            \ckey{constraints} $e_1 < s_2$
        \end{chronicle}
        \caption{\textit{\ac{UAV}} method}
        \label{fig:goto-methods-uav}
    \end{subfigure}
    \caption{Some method chronicles used to decompose the \textsf{goto} task}
    \label{fig:goto-methods}
\end{figure}

The \autoref{fig:goto-naive} shows a possible decomposition of the $[s,e]goto(v, t)$ task made by a user.
There are four possibilities for the vehicle $v$ to go to the location $t$:
\begin{itemize}
    \item It is already at the location, \ie $loc(v)=t$, then there is no operation (\textit{Noop}) to do. The associated chronicle is detailed in the \autoref{fig:goto-methods-noop}.
    \item It is a \ac{UAV}, then it flies to another location and retry to go to $t$ from this new location. The recursion will end when $loc(v)=t$ with the \textit{Noop} method. The associated chronicle is detailed in the \autoref{fig:goto-methods-uav}.
    \item In the same way as \acp{UAV}, the \acp{UGV} and humans will move and try again. The associated chronicles are similar to the one of \textit{\ac{UAV}}.
\end{itemize}

\subsection{Explore task}

\begin{figure}[htbp]
    \centering
    \scalebox{0.65}{
        \begin{forest}
  [{explore(r, f, t)}, task
      [{forward(r, f, t)}, method
          [{air(r, f, t)}, method
              [{goto(r, f)}, task]
              [{explore air(r, f, t)}, action]
          ]
          [{ground(r, f, t)}, method
              [{goto(r, f)}, task]
              [{explore ground(r, f, t)}, action]
          ]
      ]
      [{backward(r, t, f)}, method
          [{air(r, t, f)}, method
              [{goto(r, t)}, task]
              [{explore air(r, t, f)}, action]
          ]
          [{ground(r, t, f)}, method
              [{goto(r, t)}, task]
              [{explore ground(r, t, f)}, action]
          ]
      ]
  ]
\end{forest}
    }
    \caption{Natural decomposition of the \textsf{explore} task}
    \label{fig:explore-naive}
\end{figure}
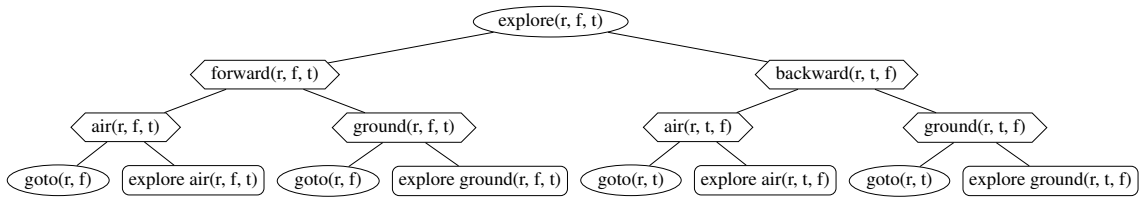

The robots need to be able to explore an edge the navigation graph in order to detect the obstacles and secure the path for the humans.
To explore the edge going from the location $f$ to the location $t$, the robot $r$ needs to be either in location $f$ or location $t$.
Therefore, there are two methods to explore an edge (shown in \autoref{fig:explore-naive}):
\begin{itemize}
    \item going to the location $f$ then explores from $f$ to $t$: \textit{forward} method
    \item going to the location $t$ then explore from $t$ to $f$: \textit{backward} method
\end{itemize}

These two methods can be accomplished by a \ac{UAV} with the \textit{air} method or by a \ac{UGV} with the \textit{ground} method.
The distinction between the two associated actions is that one effect of the \textit{explore air} action will be $explored\ air(f, t)\gets \top$, and for the \textit{explore ground} action it will be $explored\ ground(f, t)\gets \top$.
These two state variables are used in conditions of the \textit{walk} action in order for the humans to move securely.

As for the movement actions, the duration of an exploration is based on the state variable $dur(v, f, t)$.


\subsection{Secure task}

\begin{figure}[htbp]
    \centering
    \begin{forest}
  [{secure(r, o)}, task
      [{Obs$_1$(r, o)}, method
          [{goto(r, loc(o))}, task]
          [{secure Obs$_1$(r, o)}, action]
      ]
      [\dots, edge=dashed]
      [\dots, edge=dashed]
      [\dots, edge=dashed]
      [\dots, edge=dashed]
      [{Obs$_n$(r, o)}, method
          [{goto(r, loc(o))}, task]
          [{secure Obs$_n$(r, o)}, action]
      ]
  ]
\end{forest}
    \caption{Natural decomposition of the \textsf{secure} task}
    \label{fig:secure-naive}
\end{figure}
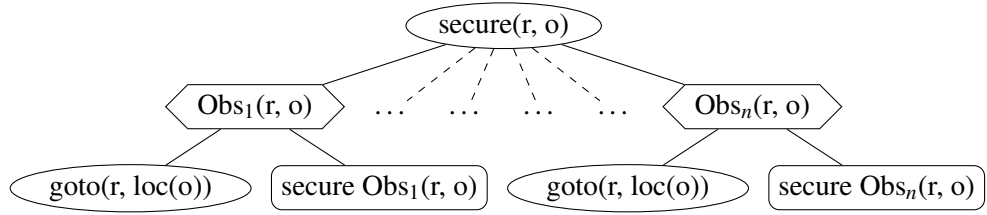

Finally, the robots need to be able to secure detected obstacles so that they can be crossed by humans.
Because there are several types of obstacles (see \autoref{fig:type-hierarchy}), there will be several methods to secure them as shown in the \autoref{fig:secure-naive}.

We made the assumption that the robot $r$ needs to be close to the obstacle $o$ to secure it for every method.
In the case where it is not needed, \eg in a military context as CoHoMa II where some obstacles representing enemy’s troops could be secured in distance with artillery fire, the associated \textit{goto} task should be removed.

For the following simulations, we consider only one way to secure an obstacle with the duration of $15$ minutes.

\subsection{Initial State and Objectives}

Once the different high-level tasks have been defined, an initial chronicle needs to be specified to encode the initial state and the objectives.

\begin{minipage}{0.5\linewidth}
    \begin{chronicle}
        \cname{$[s,e]initial$}
        \ckey{constraints} $s=0$
        \ckey{effects} $[s,s]loc(H)\gets L1$
        \cnl $[s,s]loc(UAV)\gets L1$
        \cnl $[s,s]loc(UGV)\gets L1$
        \cnl $[s,s]path(L1,L2)\gets \top$
        \cnl \vdots
        \cnl $[s,s]path(L12,L13)=\top$
        \ckey{subtasks} $[s_1, e_1]goto(H, L8)$
    \end{chronicle}
\end{minipage}
\begin{minipage}{0.4\linewidth}
    The initial chronicle starts at the timepoint $0$ and ends at the timepoint $e$.
    This timepoint can be used to specify objectives.

    Initially, the vehicles are located to the location $L_1$ and the different paths are specified.
    All the unspecified state variable values are considered to be false.

    The objective of the problem is for the humans to go to the location $L_8$.
\end{minipage}

With this initial chronicle, the planner will try to achieve the subtask $[s_1, e_1]goto(H, L8)$.
Since there are no explored paths, this is impossible without the intervention of a robot, but they cannot explore because exploration tasks are not present in the initial chronicle's subtasks.
However, the robots are not expected to explore all the paths, they are expected to be free to do whatever they want in order to help the humans.


\subsection{Freedom Task}

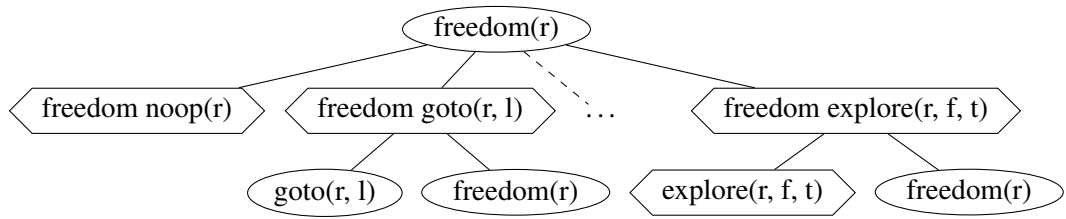
\begin{figure}[htbp]
    \centering
    \begin{forest}
  [{freedom(r)}, task
      [{freedom noop(r)}, method]
      [{freedom goto(r, l)}, method
          [{goto(r, l)}, task]
          [{freedom(r)}, task]
      ]
      [\dots, edge=dashed]
      [{freedom explore(r, f, t)}, method
          [{explore(r, f, t)}, method]
          [{freedom(r)}, task]
      ]
  ]
\end{forest}
    \caption{Natural decomposition of the \textsf{freedom} task}
    \label{fig:freedom-naive}
\end{figure}

In order to achieve that, the $freedom(v)$ task (see \autoref{fig:freedom-naive}) is added.
It allows the vehicle $v$ to go to another location or to explore a path without any constraints.
Once the robot will have nothing more to do, the $freedom\ noop(v)$ method will allow it to stop.

Next, the three subtasks $[s_2, e_2]freedom(UAV)$, $[s_3, e_3]freedom(UGV)$, and $[s_4, e_4]freedom(H)$ can be added to the initial chronicle's subtasks.
Note that the $freedom(H)$ task only allows the humans to go wherever they want, they cannot do exploration even if it is present in the decomposition of the task.
This is caused by the type hierarchy and the definition of the $explore$ task that only take robots as parameters.

In general these freedom tasks allow the planner to insert some classes of actions in the plans regardless of the rest of the hierarchy. In this sense, it simulates in the HTN the notion of task insertion~\cite{Geier2011OnTD}, where any action can be inserted along the hierarchy. It is in particular close to the \textit{task-independent} action in FAPE~\cite{BitMonnot2020FAPEAC}, where only a subset of the actions are allowed to be inserted arbitrarily.


\subsection{First Planning Results}

Considering ground truth shown in the \autoref{fig:mission-0}, the vehicles are in the location $L_1$ and the humans need to go to the location $L_8$, but there are undetected obstacles on the path.
Because the terrain is not fully known at the beginning of the mission, a replanning step is needed when the operator adds some details to the mission, \eg when an obstacle is detected by a robot.

Initially, the knowledge of the terrain is empty.
Therefore, the decision-making aid has the navigation graph shown in the \autoref{fig:mission-1} and will propose the associated plan.
This plan is to take the shortest route to the goal, with the robots ahead of the humans to secure the path.
The planning operation has been done with the Aries planner~\cite{bitmonnot:ipc:2023}, it took $333.59$s to find the optimal solution.

During the execution of that plan, the robots detect an obstacle at the location $L_5$ (see \autoref{fig:mission-2}).
A new plan is proposed based on the new knowledge of the terrain.
The planner believes it's quicker to go back and explore a new route than to secure the current obstacle.
This plan has been found in $328.25$s.

Finally, a new obstacle is discovered at the location $L_2$ (see \autoref{fig:mission-3}).
Again, a new plan is proposed based on the new knowledge of the terrain.
This time, it is faster to secure the current obstacle and go to the target.
The planner took $308.72$s to find this plan.

\begin{figure}[htb!]
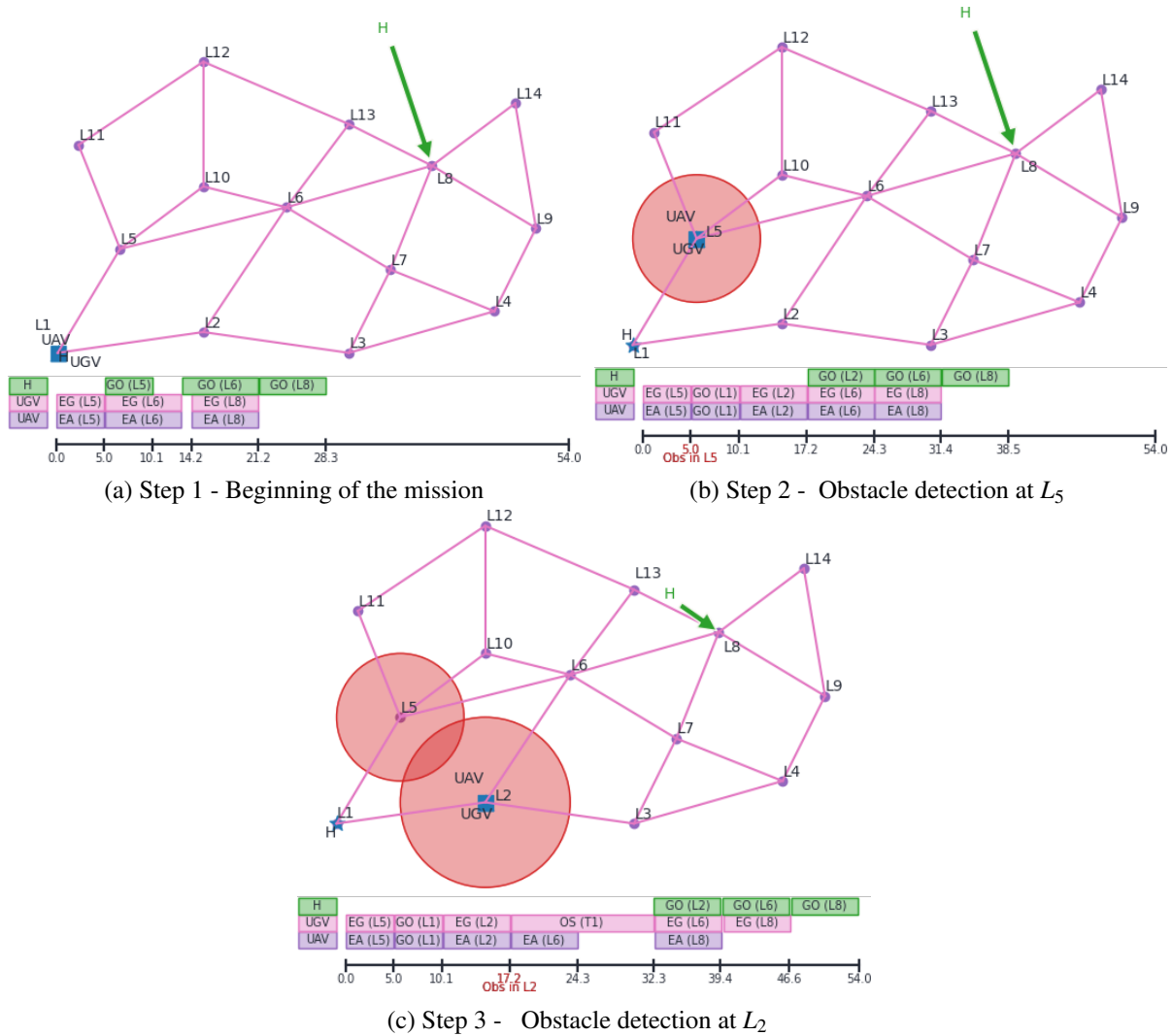

    \centering
    \foreach \n in {1, 2, 3}{
            \begin{subfigure}{0.49\linewidth}
                \centering
                \includegraphics[width=.9\linewidth]{img/mission_\n.png}
                \includegraphics[width=\linewidth]{img/plan_\n.png}
                \caption{Step \n\ -
                    \ifthenelse{\n=1}{Beginning of the mission}{
                        \ifthenelse{\n=2}{Obstacle detection at $L_5$}{
                            \ifthenelse{\n=3}{Obstacle detection at $L_2$}{}
                        }
                    }
                }
                \label{fig:mission-\n}
            \end{subfigure}
        }

    \caption{Terrain knowledge with their associated plan to solve the problem}
    \begin{tabular}{r@{: }l r@{: }l}
        $EA$ & Explore Air    & $GO$ & Goto (with $walk$, $fly$, or $roll$) \\
        $EG$ & Explore Ground & $OS$ & Secure Obstacle
    \end{tabular}
    \label{fig:mission}
\end{figure}


\section{Optimizations}

While reasonable, planning times of a handful of minutes are far from ideal in mixed-initiative planning context, especially when task durations are faster than the minute.
To reduce this time, one could ask for the first solution found by the planner (instead of an optimal solution) with the risk of handing out bad quality solutions.
Instead, in this section, we introduce some modifications that can be brought to the planning model in order to speed up the planning process.



\subsection{Recursive Tasks}

To find a solution, the planner needs to scan the search tree and prune the branches that lead to no solution.
Therefore, the model should use the least possible recursive tasks in order to reduce the size of the search tree.

Considering the $goto$ task (see \autoref{fig:goto-naive}) and $n>0$ the decomposition depth, \ie the number of times $goto$ leaves are replaced by the decomposition.
Note that if a leaf is not decomposed, the associated method is removed from the three since it will not be applicable.
Then, the size of the tree, \ie the number of nodes, is $2+3*4^n=\mathcal{O}(4^n)$ which is \textbf{exponential}.

\begin{figure}[htbp]
    \centering
    \begin{forest}
  [{goto(v, t)}, task
      [{Noop(v, t)}, method]
      [{M_Goto(v, t)}, method
          [{goto once(v, t)}, task
              [{UAV(v, f, i, t)}, method
                  [{fly(v, f, i)}, action]
              ]
              [{UGV(v, f, i, t)}, method
                  [{roll(v, f, i)}, action]]
              [{Humans(v, f, i, t)}, method
                  [{walk(v, f, i)}, action]]
          ]
          [{goto(v, t)}, task]
      ]
  ]
\end{forest}
    \caption{Optimized decomposition of the \textsf{goto} task}
    \label{fig:goto-opti}
\end{figure}
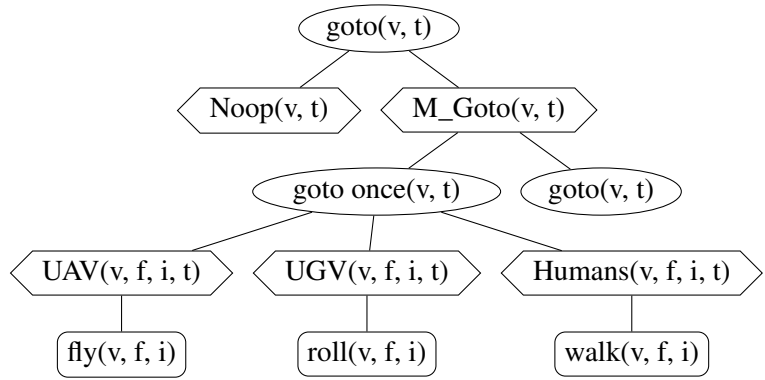

However, one can notice that all the methods have the same pattern.
There is an action followed by the recursive call to the $goto$ task.
Then, the actions can be grouped in a $goto\ once$ task and the $goto$ task can be moved outside in order to be present only once as shown in the \autoref{fig:goto-opti}.

With this new decomposition and $n>0$ the decomposition depth, the size of the tree is $2+10*n=\mathcal{O}(n)$ which is \textbf{linear}.

This method can be applied to all the tasks.
For $goto$, $explore$, and $secure$, it is the call to $goto$ which will be extracted.
For the $freedom$ task, it is the recursive call to $freedom$.


\subsection{Complete Navigation Graph}

One of the mission's assumptions is that the calculated plan is not intended directly for the robots, but for a human operator to approve, so the plan must be as simple as possible, which translates in particular into the aggregation of movement actions.

The edges of the navigation graph can be set to prohibit the passage of certain vehicles.
Therefore, one could make the graph complete, \ie each node is connected to all the others, and make an edge allowed for a given vehicle if :
\begin{itemize}
    \item the vehicle is a robot, humans need to know exactly where they are going
    \item there is a path in the initial graph corresponding to that new edge
    \item the vehicle is allowed to go through all the edges of this path
    \item the time taken by the vehicle to pass this new edge is the time taken to cover the associated path
\end{itemize}

\noindent
This way, the action of going from $L_1$ to $L_9$ for a robot can be done with only one decomposition of the $goto$ task rather than $4$ decomposition without this navigation graph manipulation.
As a result, the search tree will be smaller, and the solution will be found more quickly.


\subsection{Objectives as Conditions}

The initial chronicle defines the objective as a subtask.
That means the given subtask needs to be accomplished.
However, the subtasks also contain three $freedom$ tasks in order to the vehicles to do whatever they want to complete the objective.

Looking closer, one can see that this allows many opportunities to achieve the objective $goto(H, L8)$, which are all the possible combinations of the two tasks $goto(H, L8)$ and $freedom(H)$, both allowing the humans to move.

To avoid that, the objective can be encoded in another way.
The objective is not for the humans to go to the location $L_8$, but to be at the location $L_8$ at the end, \ie $loc(H) = L_8$.
Therefore, the initial chronicle can be updated as shown below.
With this new encoding, the only way for the humans to be at the location $L_8$ is to use the $goto(H, L8)$ hidden in the $freedom(H)$ subtask.
As a result, the search tree will be reduced.

\begin{chronicle}
    \cname{$[s,e]initial$}
    \ckey{constraints} $s=0$
    \ckey{effects} $[s,s]loc(H)\gets L1$
    \cnl $[s,s]loc(UAV)\gets L1$
    \cnl $[s,s]loc(UGV)\gets L1$
    \cnl \dots
    \ckey{\bf conditions} $[e,e]loc(H)=L8$
    \ckey{subtasks} $[s_1, e_1]freedom(H)$
    \cnl $[s_2, e_2]freedom(UAV)$
    \cnl $[s_3, e_3]freedom(UGV)$
\end{chronicle}


\subsection{Final Planning Results}

Considering the same mission studied in the previous section, the planner found the same plans as shown in the \autoref{fig:mission}.
This demonstrates that the proposed optimizations do not change the problem represented by the model, both are equivalent.
However, as shown in the \autoref{tab:time-comparison}, the planner is $95$\% faster with these optimizations.

\begin{table}[!htbp]
    \centering
    \begin{tabular}{lccc} \hline
        \textbf{}                 & \textbf{Step 1} & \textbf{Step 2} & \textbf{Step 3} \\ \hline
        \textbf{Natural model}    & $333.59$s       & $328.25$s       & $308.72$s       \\
        \textbf{Optimized model}  & $13.61$s        & $14.17$s        & $9.19$s         \\ \hline
        \textbf{Global reduction} & $95.9$\%        & $95.7$\%        & $97.0$\%        \\
    \end{tabular}
    \caption{Planning time with and without the proposed optimizations}
    \label{tab:time-comparison}
\end{table}

As these optimizations are independent of the domain, the same results should be observed in other use cases.


\section{Conclusion}


In this paper, we presented a planning-based decision-making aid that exploits a hierarchical task planner for the control of a fleet of robots in an exploration scenario.
A first natural model of the problem has been proposed. We then proposed some domain-agnostic optimization of this initial model, which resulting in the planner being at least 20 times faster to provide an optimal solution.

Some assumptions have been made in the current model, notably that the robot's battery level is infinite.
It could be interesting to be able to represent these kinds of resources in order to accomplish more complex missions.
Moreover, the planner is optimizing the makespan of the plan, \ie it tries to make the plan as short as possible in time.
It could be useful to associate a cost to each action to optimize the global cost of the plan, \ie the sum of the present action's cost.
This way, it could be possible to minimize, for example, the total distance travelled by the human group.


\bibliographystyle{eptcs}
\bibliography{main}

\listoftodos

\end{document}